\documentclass[11pt]{article}
\usepackage{algorithm}
\usepackage{algorithmic}
\usepackage{amsmath}
\usepackage{amssymb}
\usepackage{adjustbox}
\usepackage{paralist} % 添加到导言区
\usepackage{bbding}
\usepackage{pifont}
\usepackage{enumitem}
\usepackage{listings}
\usepackage{xcolor}

\lstdefinestyle{mdblock}{
  basicstyle=\ttfamily\small, % 等宽字体
  numbers=left,               % 左侧行号
  numberstyle=\tiny,
  stepnumber=1,
  numbersep=8pt,
  frame=tb,              % ✅ 只要顶部和底部边框
  framerule=0.5pt,       % 边框粗细
  breaklines=true,            % 自动换行
  columns=fullflexible,
  backgroundcolor=\color[gray]{0.96} % 淡灰底色（可去掉）
}

\usepackage{flushend}
\usepackage{multirow} 
\usepackage{booktabs}
\newcommand{\FUNCTION}[1]{\STATE \textbf{Function} #1}
\newcommand{\ENDFUNCTION}{} 
\usepackage[preprint]{acl}
\usepackage{enumitem}
\usepackage{times}
\usepackage{latexsym}
\usepackage[T1]{fontenc}
\usepackage[utf8]{inputenc}

\usepackage{microtype}

\usepackage{inconsolata}

\usepackage{graphicx}

\title{NED-Tree: Bridging the Semantic Gap with Nonlinear Element Decomposition Tree for LLM Nonlinear Optimization Modeling}

\author{
 \textbf{Zhijing Hu\textsuperscript{1}},
 \textbf{Yufan Deng\textsuperscript{1}},
 \textbf{Haoyang Liu\textsuperscript{2}},
 \textbf{Changjun Fan\textsuperscript{1}}
\\
 \textsuperscript{1}College of System Engineering, National University of Defense Technology,
 \\
 \textsuperscript{2}University of Science and Technology of China
\\
 \small{
   \textbf{Correspondence:} \href{fanchangjun09@163.com}{fanchangjun09@163.com}
 }
}

\begin{document}
\maketitle
\begin{abstract}
% This document is a supplement to the general instructions for *ACL authors. It contains instructions for using the \LaTeX{} style files for ACL conferences.
% The document itself conforms to its own specifications, and is therefore an example of what your manuscript should look like.
% These instructions should be used both for papers submitted for review and for final versions of accepted papers.
Automating the translation of Operations Research (OR) problems from natural language to executable models is a critical challenge. While Large Language Models (LLMs) have shown promise in linear tasks, they suffer from severe performance degradation in real-world nonlinear scenarios due to semantic misalignment between mathematical formulations and solver codes, as well as unstable information extraction. In this study, we introduce \textbf{NED-Tree}, a systematic framework designed to bridge the semantic gap. \textbf{NED-Tree} employs (a) a sentence-by-sentence extraction strategy to ensure robust parameter mapping and traceability; and (b) a recursive tree-based structure that adaptively decomposes complex nonlinear terms into solver-compatible sub-elements. Additionally, we present \textbf{NEXTOR}, a novel benchmark specifically designed for complex nonlinear, extensive-constraint OR problems. Experiments across 10 benchmarks demonstrate that \textbf{NED-Tree} establishes a new state-of-the-art with 72.51\% average accuracy, NED-Tree is the first framework that drives LLMs to resolve nonlinear modeling difficulties through element decomposition, achieving alignment between modeling semantics and code semantics. The NED-Tree framework and benchmark are accessible in the anonymous repository. \footnote{\url{https://anonymous.4open.science/r/NORA-NEXTOR.}}.

\end{abstract}

\section{Introduction}
\label{sec:introduction}

Operations Research (OR) plays a pivotal role in practice by addressing complex OR problems through the construction of mathematical models and the application of optimization algorithms \citep{saban2021procurement,corbett2001shared}. Its applications are widespread, spanning from reducing energy costs and optimizing supply chains to enhancing profits \citep{singh2012overview, Brandimarte_1993} and solving challenges in job scheduling \citep{Hong_Ma_Banerjee_Mao_2016} and path planning \citep{Li_Mellou_Zhang_Pathuri_Menache_2023, wang2020combinatorial}, among numerous other domains \citep{qian2021derivative,trimborn2020investing}. However, translating real-world problems described in natural language into machine-solvable mathematical models remains a long-standing and critical bottleneck \citep{JiangShu2025llmopt}. This process not only requires highly specialized expertise to accurately define variables, constraints, and objectives but also often involves iterative refinement and modification, posing a barrier for non-experts and thereby limiting the broader application of optimization techniques \citep{ghiani2022introduction,cornuejols2018optimization,rao2019engineering}.

Recent advances in LLMs have accelerated research into automating the transformation of natural language descriptions into executable optimization models. Existing work can be broadly divided into two categories: fine-tuning methods and prompt-based agents. Prompt-based agent frameworks\citep{xiao2024chainofexperts,ahmaditeshnizi2024optimus} have emerged as promising alternatives for handling complex problems. Besides, fine-tuning methods \citep{huang2025orlm,yang2024optibench,lu2025optmath, JiangShu2025llmopt} leverage large-scale instruction datasets, they suffer from prohibitive computational costs. Parallel advancements in evaluation benchmarks have progressed from early tasks \citep{ramamonjison2023nl4opt} to more comprehensive benchmarks. \citep{huang2024mamo,xiao2024chainofexperts,huang2025orlm,yang2025ORThought}

\begin{figure*}
  \centering
  \includegraphics[width=1\linewidth]{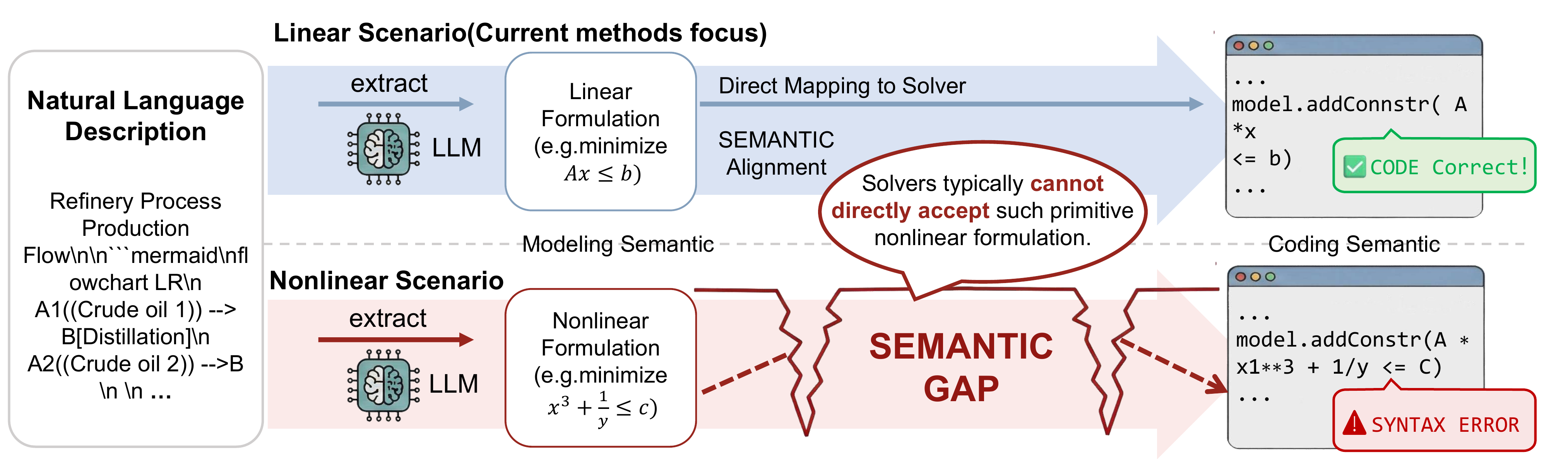}
  \caption{The Nonlinear Semantic Gap in LLM Optimization Modeling }
  \label{fig:motivation}
\end{figure*}

Despite significant progress in Linear Programming (LP), real-world OR scenarios present distinct challenges that current methods fail to address. In actual industrial environments, nonlinear terms—such as fractions, high-order powers and exponentials —are prevalent, yet existing research largely overlooks these factors. Consequently, existing methods cannot effectively model nonlinear problems. Our experiments on modeling nonlinear problem settings demonstrate a marked performance degradation when applying current methods; even minor nonlinear perturbations cause accuracy to plummet severely (see Section ~\ref{sec:LLMNOMO} for details), as illustrated in Figure~\ref{fig:motivation}. This performance collapse reveals challenges that existing methods face in nonlinear problem. First is semantic gap between modeling and coding. While nonlinear symbolic relationships (e.g., $Ax^\alpha e^x \le C$) are mathematically intuitive, mainstream solvers (such as Gurobi) typically cannot process these raw formulas directly. Instead, they require specific nonlinear API calls or equivalent reformulations via auxiliary variables. LLMs often generate ``mathematically correct'' formulas but, lacking awareness of low-level solver constraints, produce code that fails during execution due to syntax incompatibility or non-convexity errors. Moreover, the traditional ``All-in-one Extraction'' strategy proves unreliable when extracting OR problem factors, as LLMs are highly prone to information omission or hallucinations.

To address these challenges, we propose the \textbf{N}onlinear \textbf{E}lement \textbf{D}ecomposition Tree (NED-Tree) framework. This approach utilizes a ``Semantic Alignment Structure'' to eliminate the semantic gap between natural language, mathematical modeling, and solver code. The framework consists of two stages: 1) We employ LLMs for Sentence-by-Sentence Extraction, establishment of an explicit mapping from each parameter and constraint to the original text to build a reliable nonlinear element set; 2) We utilize the LLM to construct a recursive NED-Tree structure, adaptively decomposing complex nonlinear terms into a series of simpler atomic elements and accurately mapping them to function calls accepted by the solver. In summary, our contributions are as follows:
\begin{itemize}[leftmargin=*]
\item We introduce the \textbf{NED-Tree}, the first framework that drives LLMs to resolve nonlinear modeling difficulties through element decomposition, combined with the sentence-by-sentence extraction strategy, achieving alignment between modeling semantics and code semantics.

\item We construct the \textbf{NEXTOR} benchmark to improve evaluation, including long-text descriptions and complex nonlinearities closer to reality;

\item Experiments across 10 benchmarks demonstrate the effectiveness of our method. The NED-Tree not only excels in complex nonlinear tasks, but also performs outstandingly on linear datasets, achieving 72.51\% average accuracy—surpassing the best fine-tuning model by 13.02\% and the leading non-fine-tuning model by 6.27\%.
\end{itemize}
\begin{figure*}
    \centering
    \includegraphics[width=1\linewidth]{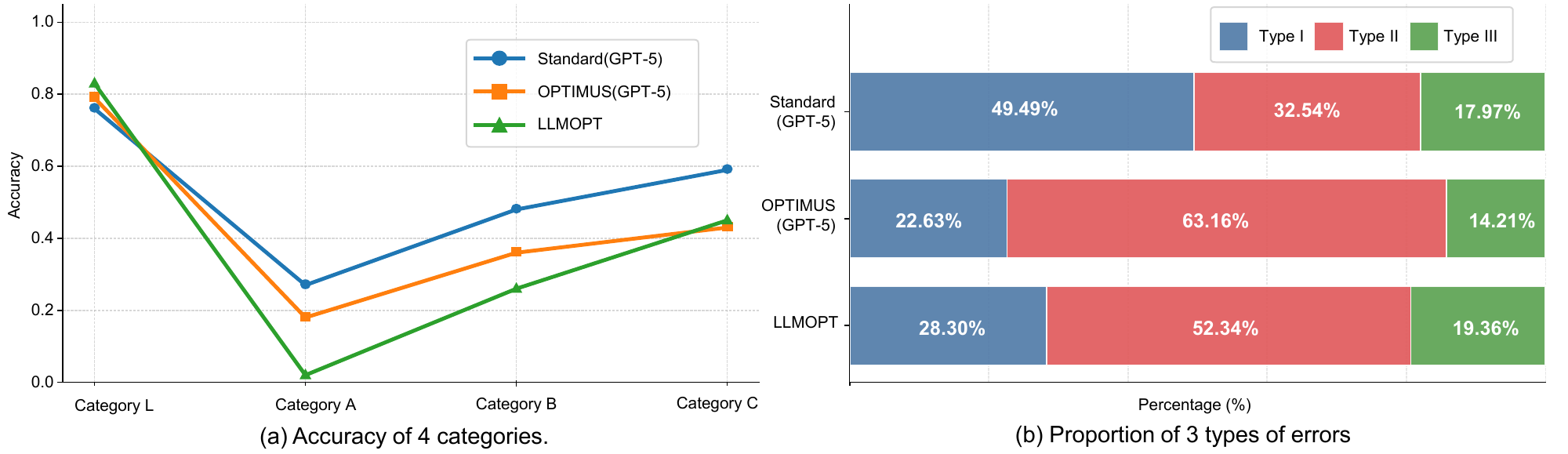}
    \caption{LLM Nonlinear Optimization Modeling in Existing Methods. (a) Accuracy of 4 categories. Category L (Linear): Linear baselines; Category A (Non-quadratic Powers): Involving high-order power terms; Category B (Fractional/Rational): Introducing complex ratios or average cost constraints; Category C (Logic/Indicator): Containing piecewise functions or conditional costs.(b) Proportion of 3 types of errors: type I: modeling semantic errors, type II: nonlinear code semantic errors, type III: other code writing errors.}
    \label{fig:observation}
\end{figure*}

\section{Related Work}
\label{RelatedWork}
\textbf{LLM Frameworks for OR.}
The emerging field of ``LLM for OR'' addresses the translation of natural language into formal optimization models through two primary technical routes. The first route employs prompt-based agent frameworks leveraging general-purpose LLMs. Early works like OPRO \citep{yang2023OPRO} pioneered using LLMs as optimizers. Subsequent frameworks such as CoE \citep{xiao2024chainofexperts} utilize multi-agent systems with iterative reflection, while OptiMUS \citep{ahmaditeshnizi2024optimus} introduces connection graphs to manage variable-constraint relationships, and OR-LLM-Agent \citep{zhang2025orllmagent} focuses on code self-repair. The second route centers on fine-tuning specialized models via high-quality data synthesis. LLaMoCo \citep{ma2024llamoco} proposes a code-to-code fine-tuning framework. Innovations in synthesis include ORLM's \citep{huang2025orlm} ``seed-expansion,'' OptiBench's \citep{yang2024optibench} ``reverse generation,'' OPTMATH's \citep{lu2025optmath} bidirectional synthesis, and LLMOPT's \citep{JiangShu2025llmopt} unified ``5-element'' paradigm. These specialized models often achieve performance comparable to larger general-purpose models on specific benchmarks.

\textbf{Evaluation Benchmarks for OR.}
The advancement of LLMs for OR parallels the evolution of evaluation benchmarks. Early datasets like LPWP \citep{ramamonjison2023nl4opt} were limited to simple linear programming. Benchmarks subsequently increased in complexity; ComplexOR \citep{xiao2024chainofexperts} and NLP4LP \citep{ahmaditeshnizi2024optimus} introduced longer descriptions and implicit constraints, while MAMO \citep{huang2024mamo} expanded the scope to Mixed-Integer Linear Programming (MILP). IndustryOR \citep{huang2025orlm} enhanced applicability by sourcing problems from industrial contexts. Recently, benchmarks have further pushed boundaries: OptiBench \citep{yang2024optibench} incorporated non-linear problems and tabular data, addressing a major gap. Furthermore, OPTMATH-Bench \citep{lu2025optmath} and ORThought \citep{yang2025ORThought} targets ``hard cases'' with complex logical structures and diverse constraints to rigorously test model robustness and reasoning capabilities.

\section{LLM Nonlinear Optimization Modeling Observations}
\label{sec:LLMNOMO}

Existing research predominantly focuses on Linear Programming (LP) solutions. Consequently, a natural and critical question arises: Do Large Language Models (LLMs) possess the capability to address nonlinear problems that are widely prevalent in real-world industrial scenarios? To investigate this, We evaluated three representative methods: Standard (GPT-5), OptiMUS(GPT-5 based) , and LLMOPT by introducing subtle nonlinear perturbations to classic linear datasets, thereby transforming them into nonlinear problems into four types. Detailed perturbation strategies and mathematical formulations are provided in Appendix \ref{app:appendix_setup}. The observation results are presented in Figure \ref{fig:observation}, and confirm three core phenomena: \textbf{(a) Significant "Performance Cliff".} As shown in Figure \ref{fig:observation}(a), while all existing methods perform excellently on Category L tasks, introducing slight nonlinear perturbations (Categories A/B/C) causes a precipitous drop 20\%-50\% range in end-to-end accuracy, indicating a heavy reliance on memorized linear patterns and a lack of generalization for nonlinear structures. \textbf{(b) Deep Misalignment between Modeling and Coding Semantics.} As illustrated in Figure \ref{fig:observation}(b), the majority of failures stem not from logical misunderstanding, but from the gap between modeling semantics (recognizing nonlinearity) and code semantics (knowing how the solver accepts nonlinear inputs). \textbf{(c) Instability in Element Extraction.} We observed that Category B or C problems frequently lead to information omission (Type I errors), such as neglecting implicit non-zero denominator constraints or hallucinating parameters in piecewise costs.
%, demonstrating that traditional "All-in-one Extraction" strategies are unreliable. 
These findings prompted us to consider how to pass semantic gap, highlighting the necessity of proposing the NED-Tree framework.
\begin{figure*}
    \centering
    \includegraphics[width=1\linewidth]{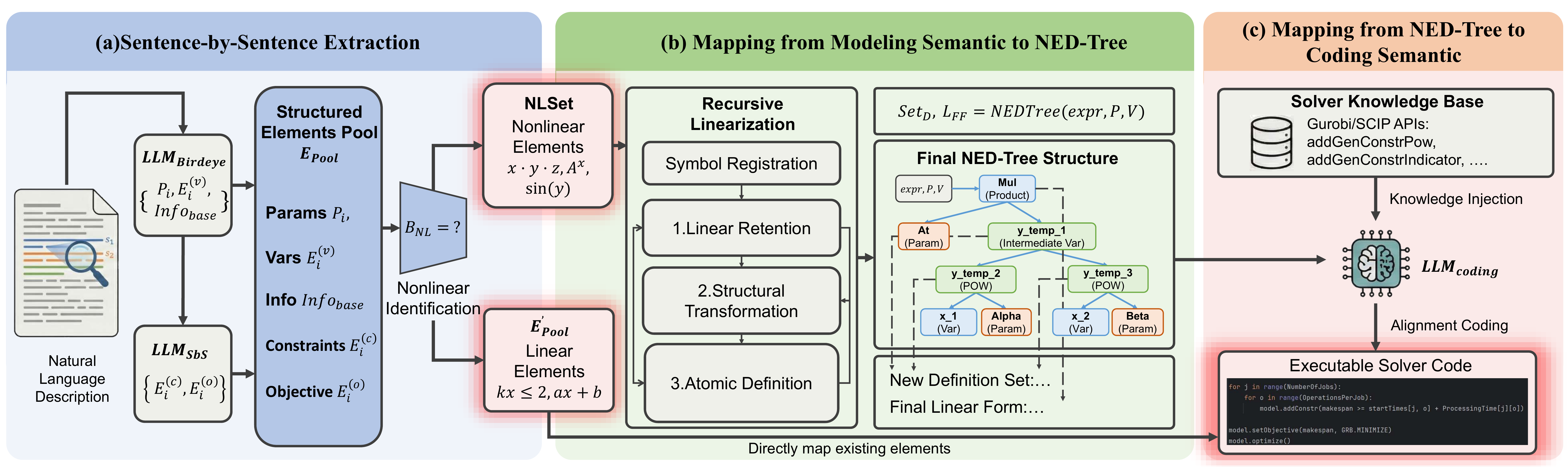}
    \caption{NEDTree framework. Our approach comprises three parts: (a) Sentence-by-Sentence Extraction, (b) Mapping from Modeling Semantic to NED-Tree, and (c) Mapping from NED-Tree to Coding Semantic. The aim is to align modeling semantics with code semantics.}
    \label{fig:S4_framework}
\end{figure*}

\section{NED-Tree: Sentence-by-Sentence Extraction and Nonlinear Element Decomposition}
\label{sec:ned_tree}

To address the ``Nonlinear Performance Cliff'' observed in Section~\ref{sec:LLMNOMO} and the deep misalignment between natural language modeling semantics and solver code semantics, this chapter proposes the NED-Tree framework. This framework no longer relies on black-box end-to-end generation but constructs an interpretable intermediate representation layer. By logically deconstructing complex nonlinear terms, registering symbols, and performing recursive transformations, it ensures that the mathematical models generated by LLMs can be accurately mapped into atomic operation sequences executable by solvers, thereby fundamentally eliminating the semantic gap.

\subsection{Preliminary}
\label{sec:preliminary}

\subsubsection{OR Problem Automated Modeling and Solving Definition}
\label{sec:or_definition}

We define the automated modeling and solving of Operations Research (OR) problems as an end-to-end mapping task from natural language to executable code. Given an OR problem $p$ described in natural language, the modeling method $\Theta(\cdot)$ first constructs its formal mathematical model, then converts it into executable Python code (such as calling Gurobi or SCIP interfaces). Under the assumption that the solver is reliable, the executed code produces a numerical solution $ans$, denoted as
\begin{equation}
    ans = \Theta(p)
\end{equation}
This paper focuses on complex nonlinear problems $p_{NL}$, which cover Mixed-Integer Nonlinear Programming (MINLP), including but not limited to High-Order Power (HP), Fractions and Fractional Powers (FP), and Exponential and Logarithmic (ELP) nonlinear terms. Their general form $t_{NL}$ can be represented as
\begin{equation}
p_{NL} \sim t_{NL} = \frac{F_{NL}(X)}{G_{NL}(X)}
\end{equation}
where $F_{NL}(X)$ and $G_{NL}(X)$ are functions containing nonlinear structures. Compared to linear problems, such problems require the model not only to understand mathematical logic but also to handle non-convexity and specific constraints of solver APIs.

\subsubsection{NLSet and Nonlinear Sub-Element Definition}
\label{sec:nlset_definition}

To achieve precise semantic alignment, we define a Nonlinear Sub-Element as the smallest mappable nonlinear unit in the model. Specifically, for a term $t$ in the objective function $o$ or constraint set $C$, if it has no coefficient and its second-order partial derivative is non-zero, i.e., $\exists j, \frac{\partial^2 t}{\partial x_j^2} \neq 0$ (e.g., $x^2$ or $\cos x$), it is identified as a nonlinear sub-element. Identifying these elements is the foundation for constructing the NED-Tree, aiming to decompose complex nested terms that are difficult to encode directly (such as $e^{x^2+y}$) into atomic calls supported by the solver. Based on this, we define the nonlinear set $NLSet = \{ t \mid t \in o \cup \bigcup_i c_i(x) \}$. This set contains all nonlinear structural functions or relations to be processed, which can be single nonlinear operators or composite expressions containing nonlinear terms. The goal of the NED-Tree is to recursively transform each element in the $NLSet$ into a combination of a linear skeleton and atomic nonlinear definitions.

\subsection{Alignment for Modeling Semantic and Coding Semantic}
\label{sec:alignment}

This section details how the NED-Tree protocol establishes a deep alignment mechanism among natural language, mathematical models, and the code layer. This process is divided into three stages: sentence-by-sentence element extraction, recursive mapping from modeling semantics to the NED-Tree, and final generation from the NED-Tree to coding semantics, as shown in Figure ~\ref{fig:S4_framework}. 

\subsubsection{Sentence-by-Sentence Extraction}
\label{sec:sbs_extraction}

When processing long-text OR problems, LLMs often miss key parameters or hallucinate due to scattered attention. To overcome this defect, we introduce the Sentence-by-Sentence Extraction (SbS) strategy, as shown in Figure~\ref{fig:S4_framework}(a). Specifically, for a given sentence $s_i$ in problem $p$, we design a bird's-eye extraction strategy $LLM_{\text{Birdeye}}$ and a sentence-by-sentence extraction strategy $LLM_{\text{SbS}}$. This process first extracts from the entire text $p=\{s_1,...,s_I\}$, driving LLMs to generate a bird's-eye view extraction, identifying basic information ${Info}_{base}$ such as problem parameters $P_i$, variables $E_i^{(v)}$, and the optimization problem type, i.e.,
\begin{equation}
    \left\{ P_i, \  E_i^{(v)}, {Info}_{base} \right\} = LLM_{\text{Birdeye}}({p})
\end{equation}

Subsequently, the LLM is driven to perform fine-grained scanning tasks, primarily extracting constraints $E_i^{(c)}$ and objective functions $E_i^{(o)}$ to complete the process:
\begin{equation}
\left\{ E_i^{(c)}, E_i^{(o)} \right\} = LLM_{\text{SbS}}(s_i)
\end{equation}
Finally, the Structured Elements Pool $E_{Pool}$ is initialized by aggregating the extraction results of all sentences to obtain $NLSet$ and refresh the Structured Elements Pool to get $E^{'}_{Pool}$:
\begin{equation}
E_{Pool} = \bigcup_{i=1}^I \left\{ E_i^{(v)}, E_i^{(c)}, E_i^{(o)},P_i \right\} 
\end{equation}
\begin{equation}
E_{Pool} \xrightarrow{\text{Aggr}} \{NLSet, E^{'}_{Pool}\}
\end{equation}

During extraction, sentences with the specific form IF $F(X) \le 0$ THEN $G(X)$ are also identified as part of the constraints. The extraction module formalizes this as constraint $c_{\text{cond}}$ and adds it to $E_{Pool}$:
\begin{equation}
c_{cond, i} \sim  \mathbb{I}(F(X) \le 0) \implies G(X) \le 0  
\end{equation}
\begin{equation}
c_{cond, i}= LLM_{SbS}(s_i) \xrightarrow{}E_{Pool} 
\end{equation}

These conditional constraints will subsequently be standardized through Indicator Constraints during the modeling phase. Furthermore, Nonlinear Pre-identification extracts nonlinear mathematical relations $t$ during the final aggregation stage mentioned above, classifying them into the nonlinear element set $\text{NLSet}$ and generating a Boolean indicator value $B_{\textbf{NL}}$:
\begin{equation}
    B_{\rm \textbf{NL}} =  
        \begin{cases}
            \text{True}, & \text{if } \exists t \in E \text{ s.t. } t \in \text{NLSet} \\
            \text{False}, & \text{otherwise}
        \end{cases}
\end{equation}

If $B_{\textbf{NL}}=\text{True}$, the system activates the subsequent NED-Tree conversion flow; otherwise, it proceeds directly to the linear solving path.

\subsubsection{Mapping from Modeling Semantic to NED-Tree}
\label{sec:mapping_ned_tree}

After obtaining the nonlinear element set $NLSet$, the core challenge lies in translating human-readable mathematical formulas into machine-solvable forms. Traditional direct translation methods often fail by ignoring solver limitations. Therefore, we introduce the NED-Tree, which reconstructs the original mathematical expression tree into a separated form of a linear backbone and atomic nonlinear definitions through recursive construction, as shown in Figure~\ref{fig:S4_framework}(b). The algorithm first performs strict Symbol Registration. The system traverses $NLSet$ and classifies symbols into a parameter set $\mathcal{P}$ (constant coefficients) and a variable set $\mathcal{V}$ (optimization subjects) based on context. The preprocessing module synchronously completes LaTeX format cleaning and the explicitation of implicit multiplication, ensuring the input $expr$ is a standard symbolic expression tree. The input for the Recursive Linearization \& Tree Building process is the original expression node $expr$ along with the registered parameter set $\mathcal{P}$ and variable set $\mathcal{V}$, while the output consists of the structured definition set $Set_\text{D}$ and the final linear form $L_{\text{FF}}$:
\begin{equation}
    Set_\text{D}, \  L_{\text{FF}} =  \text{NEDTree}(expr, \mathcal{P}, \mathcal{V})
\end{equation}

Specifically, we design a Bottom-Up recursive construction algorithm. The algorithm traverses each node of the expression tree, first recursively processing all child nodes, and then determining its linear property based on the nature of the current operator:

\textbf{(a) Linear Retention}: If the operator of the current node and its child node combination satisfy linear conditions (such as addition or constant multiplication), the original structure is retained and passed upwards directly; 

\textbf{(b) Structural Transformation}: If a specific non-convex structure (such as chain multiplication or parametric base power) is detected, a transformation function is called first to map it to a solver-compatible form (such as logarithmic transformation);

\textbf{(c) Atomic Definition}: For irreducible nonlinear terms (such as $\sin, \exp$), the algorithm generates an auxiliary variable $y_{temp}$ to replace the subtree and registers the original relation into $Set_\text{D}$. This flow ensures that the returned $L_{\text{FF}}$ always maintains a linear structure, with specific logic,the algorithm  shown in Appendix~\ref{app:appendix}.

To adapt to the API characteristics of mainstream solvers like Gurobi, we embed targeted Isomorphic Structural Transformation logic within the recursive process to ensure the generated mathematical model meets the solver's input specifications while maintaining semantic equivalence.

\textbf{(a) Chain Multiplication Decomposition}. Addressing the limitation that most solvers only support quadratic constraints, for high-order chain multiplication terms (e.g., $x \cdot y \cdot z$), the algorithm does not generate a ternary product directly but recursively decomposes it into a chain of binary products. Specifically, it first generates $y_{temp\_1} = x \cdot y$, followed by $y_{temp\_2} = y_{temp\_1} \cdot z$, reducing high-order nonlinearity into multiple second-order nonlinear constraints. 

\textbf{(b) Parametric-Base Power Transformation}. Addressing the difficulty solvers typically face in directly processing $A^x$ (where $A$ is a constant parameter and $x$ is a decision variable) forms, the algorithm incorporates logarithmic identity transformation logic. When a \texttt{pow(Param, Var)} structure is detected, the system automatically applies the transformation $A^x = e^{x \ln A}$. This process is decomposed into three atomic steps within the NED-Tree: calculating the constant coefficient $c = \ln(A)$; constructing the linear intermediate variable $y_{inner} = c \cdot x$ (recursively atomized if $x$ is a complex expression); and introducing the final auxiliary variable $y_{final} = \exp(y_{inner})$. This mechanism ensures that even specific non-convex forms of power functions can be converted into the solver's general constraint exponential interface (General Constraint Exp).

\subsubsection{Mapping from NED-Tree to Coding Semantic}
\label{sec:mapping_coding}

The final stage of modeling involves precisely mapping the constructed NED-Tree into solver code, as shown in Figure ~\ref{fig:S4_framework}(c). Since the NED-Tree has already decomposed complex logic into standard atomic definitions, we only need to traverse the generated $Set_\text{D}$ and $L_{\text{FF}}$. When generating constraints and objective functions, for the linear part, we directly map existing elements to \texttt{model.addConstr}; for nonlinear atomic definitions, we call specific General Constraint APIs based on node types, as defined by different solver documentation. Furthermore, considering that interfaces differ across solvers, we inject Solver Knowledge at this stage, though the overall logic remains similar. For example, in the GUROBI solver, power operation nodes are mapped to \texttt{model.addGenConstrPow}, exponential nodes to \texttt{model.addGenConstrExp}, and indicator function logic to \texttt{model.addGenConstrIndicator}.

\begin{table*}[h]
\centering
\caption{Comparison of Accuracy (pass@1) metric between NEDTree and other methods. \underline{Underlined} results indicate results that are second only to the SOTA, while \textbf{bold} results indicate the current SOTA. $\dagger$: $\dagger$Results are cited from their original papers. $\ddagger$: ComplexOR's GT answer is based on LLMOPT's open source repository\cite{JiangShu2025llmopt}. OPTIBench and OPTMATH's best models have no weights released. IR(Improvement Rate): + indicates that NEDTree outperforms the best performing method in this category, and - indicates not. IR(with NFT) is based on the best of all non-fine-tuning methods. IR(with FT) is based on the best accuracy of all fine-tuning methods. }
\begin{adjustbox}{max width=\textwidth}
\begin{tabular}{llccccccccccc}
\toprule
\textbf{Types} & \textbf{Method} & \textbf{NLP4LP} & \textbf{NL4OPT} & \textbf{ComplexOR$\ddagger$} & \textbf{MAMO(Easy)} & \textbf{MAMO(Complex)} & \textbf{IndustryOR} & \textbf{OptMATH} & \textbf{OPTIBENCH} & \textbf{OR-llm-Agent} & \textbf{NEXTOR}  & \textbf{AVG} \\
\midrule
\multirow{2}{*}{Non-reasoning} 
  & Deepseek-V3 & 73.96\% & 87.82\% & 66.67\% & 88.30\% & 40.28\% & 33.00\% & 32.60\% & 64.79\% & 61.44\% & 26.31\% & 57.52\% \\
  & GPT-4o      & 76.03\% & 81.30\% & 66.67\% & 88.19\% & 33.17\% & 34.00\% & 34.94\% & 58.51\% & 40.96\% & 18.42\% & 53.22\% \\
\midrule
\multirow{2}{*}{Reasoning} 
  & Deepseek-R1   & 76.44\% & 90.00\% & 66.67\% & 82.97\% & 47.39\% & 33.00\% & 40.96\% & \underline{74.21\%} & 68.67\% & 40.79\% & 62.11\%\\
  & GPT-o4-mini   & 76.85\% & 88.26\% & 72.22\% & 89.26\% & 43.12\% & 34.00\% & \textbf{45.18\%} & 64.13\% & 72.29\% & \underline{52.63\%} & 63.79\% \\
\midrule
\multirow{3}{*}{Prompt-based} 
  & Chain-of-Experts & $\dagger$53.1\% & $\dagger$64.2\% & $\dagger$25.9\% & 85.28\% & 42.18\% & 33.00\% & 39.7\% & 69.26\% & 67.47\% & 51.32\% & 53.15\%\\
  & Optimus          & $\dagger$72.0\% & $\dagger$78.8\% & $\dagger$66.67\% &  89.11\% & 44.08\%  & 32.00\% & 19.28\% & 64.96\% & 48.19\% & 18.42\%  & 53.35\% \\
  & ORllmAgent       & \underline{81.40}\% & 88.26\% & 66.67\% & 89.20\% & 47.86\% & 35.00\% & \textbf{45.18\%} & 69.75\% & \underline{74.70\%} & \underline{52.63\%}  & 65.07\%\\
\midrule
\multirow{4}{*}{Fine-tuning} 
  & ORLM(best)  & 68.60\% & $\dagger$86.50\% & 55.56\% & $\dagger$82.30\% & $\dagger$37.40\% & $\dagger$38.00\% & 9.03\% & 54.55\% & 12.05\% & 11.84\% & 45.58\% \\
  & LLMOPT      & $\dagger$\textbf{83.80\%} & $\dagger$93.00\% & $\dagger$\underline{72.70}\% & $\dagger$\textbf{97.00\%} & $\dagger$\textbf{68.00\%} & $\dagger$\textbf{46.00\%} & $\dagger$40.00\% & $\dagger$66.44\% & 13.25\% & 2.63\% & 58.28\% \\

  %   & LLMOPT      & $\dagger$\textbf{83.80\%} & $\dagger$93.00\% & $\dagger$\underline{72.70}\% & $\dagger$\textbf{97.00\%} & $\dagger$\textbf{68.00\%} & $\dagger$\textbf{46.00\%} & $\dagger$40.00\% & $\dagger$66.44\% & - & 2.6\% & 70.87 \% \\

  & OPTBENCH    & - & - & - & - & - & - & - & $\dagger$66.10\% & - & - & 66.10\%\\
  & OptMATH     & - & $\dagger$\textbf{95.90}\% & - & $\dagger$89.90\% & $\dagger$\underline{54.10}\% & - & $\dagger$34.70\% & - & - & -& 68.65\% \\
\midrule
\textbf{Context-based} 
  & NEDTree & 80.16\% & \underline{94.35}\% & \textbf{83.33\%} & \underline{90.49\%} & 49.52\% & \underline{40.00\% }& \underline{42.16\%} & \textbf{86.86\%} & \textbf{81.93\%} & \textbf{76.31\%} & \textbf{72.51\%} \\
% \textbf{IR* } 
%   &   NEDTree    & -1.24\% & +6.09\% & +16.66\% & +1.29\% & +1.66\% & +5.00\% & -3.02\% & +12.65\% & +7.23\% & +23.68\% & +6.27\%\\
% \textbf{IR** } 
%   &   NEDTree    & -3.64\% & -1.55\% & +10.63\% & -6.51\% & -18.48\% & -6.00\% & +2.16\% & +20.42\% & +68.68\% & +64.47\% & +13.02\%\\

\textbf{IR(with NFT) } 
  &   NEDTree    & -   & +6.09\%   & +16.66\% & +1.29\%    & +1.66\%   & +5.00\%   & -         & +12.65\% & +7.23\% & +23.68\% & +6.27\%\\
\textbf{IR(with FT) } 
  &   NEDTree    & -   & -         & +10.63\% & -          & -18.48\%  & -         & +2.16\%   & +20.42\% & +68.68\% & +64.47\% & +13.02\%\\

\bottomrule
\end{tabular}
\label{table:ALL}
\end{adjustbox}
\begin{flushleft}
% \footnotesize{
% $\dagger$: Results reported in recent original papers. $\ddagger$: ComplexOR data is from LLMOPT's paper. \\ IR(Improvement Rate)*: IR* is based on the best of all no fine-tuning methods. IR**: IR** is based on the best accuracy of all fine-tuning methods. 
% }
\end{flushleft}
\end{table*}

\section{Experiments}
\label{sec:experiments}

\subsection{Experimental Setup}
\label{sec:setup}

In this section, we introduce the experimental design, results, and analysis. All experiments are conducted on a single GPU server equipped with an NVIDIA GeForce RTX 5090 GPU 32G. For all LLM-based methods, we utilized their APIs for inference and set a uniform temperature of 0.2 and a maximum token count of 8192. We write the code in Python 3.12 and use GUROBI 12.0.2 as the solver. We conduct 10 repeated experiments using a fixed seed of 42. Due to space limitations, we have omitted the variance in the main text. This does not mean we only conducted one experiment; we have provided complete data in the appendix.

\textbf{Benchmarks.} Our evaluation covers 10 public Operations Research benchmarks, spanning a wide range of difficulty. These problems range from foundational linear programming problems (such as NL4OPT, NLP4LP, MAMO) to industrial-grade problems with complex constraints (such as IndustryOR, ComplexOR). Crucially, to assess the capability of handling nonlinear semantics, we include challenging tasks involving highly nonlinear mathematical expressions (such as OPTIBENCH, OptMATH) as well as our newly constructed \textbf{NEXTOR} benchmark. NEXTOR is designed to be comprehensive, including challenging linear programming, various nonlinear optimization problems, and instances injected with real-text redundancy. The comparison between NEXTOR and other benchmark, generation methods and data statistics are provided in the Appendix~\ref{app:NEXTOR_Details}

\textbf{Baselines.} We compare our method with four categories of representative methods: (a) Reasoning-free Models: Deepseek-V3 and GPT-4o; (b) Reasoning Models: Deepseek-R1 and GPT-5; (c) Prompt-based Methods: Chain of Experts, OPTIMUS, and OR-LLM-Agent; (d) Fine-tuning Methods: ORLM, OPTBENCH, OptMATH, and LLMOPT.

\textbf{Settings and Metrics.} We utilize two metrics: \textbf{Accuracy (pass@1)} measures the success of end-to-end modeling and solving, and \textbf{Pass Rate (PR)} evaluates the syntactic correctness and executability of the generated code.

\subsection{Main Results}
\label{sec:main_results}

\textbf{Our Method Demonstrates Superior Overall Performance.} We first analyze the overall performance across all 10 benchmarks. As shown in Table \ref{table:ALL}, our method establishes a new state-of-the-art with a 72.51\% average accuracy.

Although specialized fine-tuned models like LLMOPT score high on the specific datasets they were trained on (e.g., NL4OPT), they struggle significantly with novel data distributions. For example, in the comprehensive OPTIBENCH dataset, our method significantly leads with an accuracy of 86.86\%, whereas fine-tuned models often exhibit the "seesaw problem"—improving on one task while degrading on others. Overall, our method achieves an average improvement of 13.02\% compared to fine-tuned baselines. Compared to non-fine-tuned baselines (prompt-based and reasoning models), our method improves by an average of 6.27\%. This result challenges the traditional notion that performance improvements in operations research modeling must rely on expensive fine-tuning. In summary, our method demonstrates strong robustness and generalization across both linear and nonlinear domains without parameter updates.

\begin{table}[htbp]
\centering
\caption{Comparison of AC and PR metrics between linear and nonlinear tasks on NEXTOR Benchmark.}
\begin{adjustbox}{max width=0.48\textwidth}
\begin{tabular}{lccccc}
\toprule

\multirow{2}{*}{\textbf{Method}} & \multicolumn{2}{c}{\textbf{Nonlinear}} & \multicolumn{2}{c}{\textbf{Linear}} \\%&& \multirow{2}{*}{\textbf{AvgAC}} \\
\cmidrule(lr){2-3} \cmidrule(lr){4-5}
& \textbf{AC} & \textbf{PR} & \textbf{AC} & \textbf{PR} \\%& \\
\midrule
Deepseek-V3         & 23.68\% & 31.58\% & 28.95\% & 71.05\% \\%& 26.31\% \\
GPT-4o              & 21.05\% & 26.32\% & 15.79\% & 52.63\% \\%& 18.42\% \\
Deepseek-R1         & 39.47\% & 55.26\% & 42.11\% & 92.11\% \\%& 40.79\% \\
GPT-5         & 60.53\% & 89.47\% & 44.74\% & 89.47\% \\%& 52.63\% \\
Chain-of-Experts    & 52.63\% & 76.32\% & 50.00\% & 92.11\% \\%& 51.32\% \\
ORLLMAgent          & 60.53\% & 81.58\% & 44.74\% & 89.47\% \\%& 52.63\% \\
LLMOPT              &  0.00\% & 13.16\% &  2.63\% & 10.53\% \\%&  2.63\% \\
ORLM                & 21.05\% & 52.63\% &  2.63\% & 44.74\% \\%& 11.84\% \\
\midrule
NEDTree & \textbf{92.11\%} & \textbf{100.00\%} & \textbf{60.53\%} & \textbf{100.00\%} \\%& \textbf{76.31\%} \\
\bottomrule
\end{tabular}
\label{table:NEXTOR}
\end{adjustbox}
\end{table}

\begin{figure*}
    \centering
    \includegraphics[width=1\linewidth]{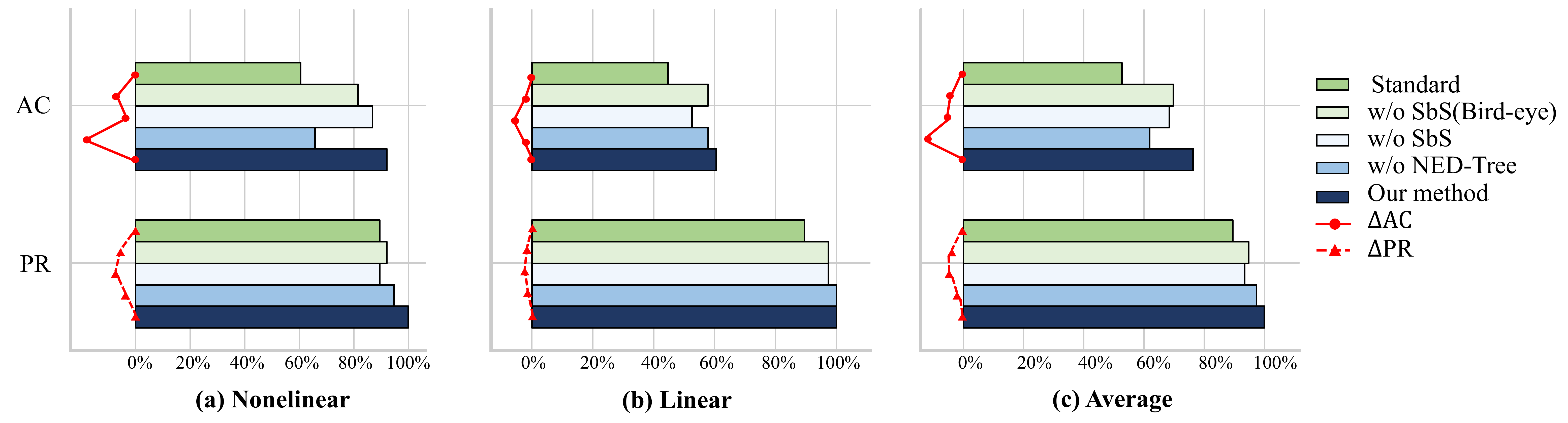}
    \caption{Ablation study of NEDTree on NEXTOR benchmark. The left line is a line graph showing the difference between the module in the ablation study and NEDTree. The further to the left, the greater the difference.}
    \label{fig:ablation}
\end{figure*}

\textbf{Our Method Effectively Handles Nonlinear Semantic Gaps.} In Section~\ref{sec:LLMNOMO}, we observed a "performance cliff" due to the semantic gap between mathematical modeling and solver API requirements. We conducted a detailed evaluation on the linear and nonlinear subtasks of NEXTOR. The results are shown in Table \ref{table:NEXTOR}. On nonlinear problems, baselines like GPT-4o and Deepseek-V3 achieved accuracies of only 21.05\% and 23.68\% respectively, while our model achieved a remarkable 92.11\% accuracy, effectively overcoming the performance cliff. Notably, across a large number of nonlinear tasks, our method achieved a Code Pass Rate (PR) of 100.00\%. This indicates that the NED-Tree structure successfully converts complex nonlinear terms (e.g., high-order powers, fractions) into solver-compatible atomic definitions, effectively eliminating the "Linear Code Semantic Error" (Error Type 2) mentioned in Section~\ref{sec:LLMNOMO}. Furthermore, while fine-tuned models like LLMOPT collapsed to nearly 0\% on NEXTOR nonlinear tasks, NED-Tree maintained high performance. This confirms that the proposed decomposition strategy effectively aligns modeling semantics with coding semantics, enabling the model to "understand" and "construct" nonlinear constraints rather than merely memorizing patterns.

\subsection{Ablation Study}
\label{sec:ablation}

To verify the contribution of each component within the model framework, we conducted an ablation study on the NEXTOR benchmark, as shown in Figure \ref{fig:ablation}. The results are summarized as follows:

\textbf{w/o SbS:} Removing the SbS strategy (Bird's-eye and Sentence-by-Sentence extraction) resulted in accuracy drops of 6.58\% and 7.89\%, respectively. This confirms that for complex long-text problems, one-pass extraction and a bird's-eye view are insufficient, and step-by-step extraction is crucial for understanding modeling semantics.

\textbf{w/o NED-Tree:} This is the most critical component for nonlinear tasks. Removing the AlignModel agent caused the accuracy on nonlinear tasks to plummet by \textbf{26.32\%}. This empirically proves that without tree-based decomposition, LLMs cannot bridge the semantic gap of complex mathematical expressions.

\subsection{Case study}

We present a case study on the decomposition of a nonlinear term common in industrial constraints: $At \cdot x_1^\alpha \cdot x_2^\beta$, and we get
`y\_temp\_1 = pow(x\_1, alpha)`, `y\_temp\_2 = pow(x\_2, beta)` and `At * y\_temp\_3`, this allows the code agent to strictly call `model.addGenConstrPow` to map the leaves and standard quadratic constraints, ensuring executability, more details can be seen in the appendix~\ref{app:Full_case_study}.

\section{Conclusion}

In this paper, we introduced NED-Tree, a systematic framework designed to bridge the critical semantic gap in nonlinear optimization modeling. By utilizing a Sentence-by-Sentence extraction strategy combined with a recursive nonlinear element decomposition mechanism, NED-Tree effectively addresses the instability of element extraction and the deep misalignment between mathematical formulations and solver codes, bypassing the need for expensive fine-tuning. To validate our approach, we presented the NEXTOR benchmark and demonstrated NED-Tree's superiority through extensive experiments. NED-Tree establishes a new state-of-the-art, achieving an average accuracy of 72.51\% across 10 benchmarks—surpassing the best fine-tuning model by 13.02\% and the leading non-fine-tuning model by 6.27\%. Ultimately, NED-Tree offers a novel and effective path for solving complex nonlinear OR problems, demonstrating that structural semantic alignment can achieve robust generalization without the need for resource-intensive fine-tuning.

\section*{Limitations}

Despite the advancements achieved by NED-Tree, several limitations remain. First, while the Sentence-by-Sentence extraction strategy improves accuracy, the information extraction mechanism can be further refined; future work aims to establish a multi-level extraction hierarchy to bolster robustness and traceability in highly ambiguous contexts. Second, the framework currently lacks a dynamic error-correction loop to address post-modeling diagnostic feedback or real-time problem shifts. 

\bibliography{custom}

\appendix

\section{Full Experimental Setup and Nonlinear Perturbation Strategy}
\label{app:appendix_setup}

We selected 100 linear Operations Research (OR) problems of simple and general difficulty from classic benchmarks such as IndustryOR, LogiOR \citep{yang2025ORThought}, and NLP4LP. By manually injecting specific nonlinear terms while preserving the core logic, we constructed a sampled dataset of 300 instances categorized into four types:Linear, Non-quadratic Powers, Fractional\/Rational and Logic/Indicator.
\begin{itemize}
    \item \textbf{Category L (Linear)}: 100 baseline linear problems;
    \item \textbf{Category A (Non-quadratic Powers)}: 94 instances where linear terms in objectives or constraints are upgraded to powers, such as modifying cost terms from $3x$ to $3x^2$ (convex quadratic) or $3x^3$ (requiring auxiliary variables);
    \item \textbf{Category B (Fractional/Rational)}: 50 instances introducing complex ratios (e.g., efficiency changed from $x+y$ to $x/y$) or average cost constraints (e.g., $\frac{\sum cost_i x_i}{\sum x_i} \le \bar{c}$), which mathematically require explicit handling of denominators like $\frac{F(x)}{G(x)}\le k$ and $G(x)\neq 0$;
    \item(d) \textbf{Category C (Logic/Indicator)}: 56 instances involving piecewise functions or conditional costs (e.g., fixed cost triggered if $x > 50$), necessitating the correct invocation of indicator constraints rather than mechanical linear superposition. We conducted 10 repeated experiments using three representative methods: the inference-based Standard (GPT-5), the prompt-framework-based OptiMUS, and the fine-tuned specialized model LLMOPT.
\end{itemize}

\section{NED-Tree Construction Algorithm}
\label{app:appendix}

The full NED-Tree Construction Algorithm is:
\begin{algorithm}[h]
\caption{NED-Tree Construction Algorithm}
\begin{algorithmic}[1]
\REQUIRE Root Expression $expr$, Parameter Set $\mathcal{P}$, Variable Set $\mathcal{V}$
\ENSURE Definition Set $Set_{\text{Definitions}}$, Linear Form $L_{\text{FinalForm}}$

\STATE Initialize $Set_{\text{Definitions}} \leftarrow \emptyset$
\STATE $L_{\text{FinalForm}} \leftarrow \text{RecursiveBuild}(expr)$
\RETURN $Set_{\text{Definitions}}, L_{\text{FinalForm}}$

\FUNCTION{RecursiveBuild($node$)}
    \STATE \textbf{Base Case:} \IF{IsAtomic($node, \mathcal{P}, \mathcal{V}$)} \RETURN $node$ \ENDIF
    
    \STATE $children' \leftarrow$ [RecursiveBuild($c$) \textbf{for} $c$ in $node.children$]
    
    \IF{IsLinear($node.op, children'$)}
        \RETURN ConstructNode($node.op, children'$)
    \ELSE
        \STATE $node_{trans} \leftarrow \text{ST(}node.op, children')$  
        \RETURN RegisterDefinition($node_{trans}, Set_{\text{D}}$)
    \ENDIF
\ENDFUNCTION

\FUNCTION{RegisterDefinition($node, D$)}
    \STATE $y_{new} \leftarrow \text{NewSymbol}("y_{temp}") $  
    \STATE $D.add(y_{new} = node)$
    \RETURN $y_{new}$
\ENDFUNCTION
\end{algorithmic}
\label{alg:ned_tree_construction}
\end{algorithm}

\section{NEXTOR Details. }
 \label{app:NEXTOR_Details}

\begin{figure*}[h]
  \centering
  \includegraphics[width=\linewidth]{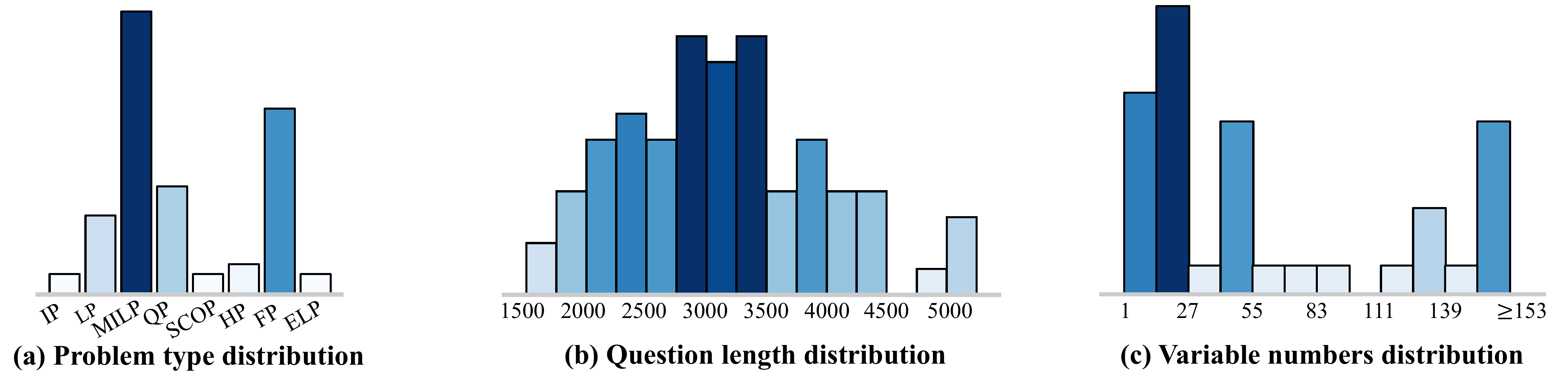}
  \caption{NEXTOR’s Statistics. (a) Problem type distribution, where HP, FP and ELP are nonlinear programming involving high-order powers (greater than 2), fractions \& fractional powers, and exponentials \& logarithms, respectively. (b) Question length distribution. (c) Variable numbers distribution.  }
    \label{fig:benchmark_Statistics}
\end{figure*}

\begin{table*}[t]
% \small
\centering
\caption{Comparison of NEXTOR with other benchmarks. HP, FP, and ELP are nonlinear programming involving high-order powers (greater than 2), fractions \& fractional powers, and exponentials \& logarithms, respectively. \ding{52} indicates coverage of this aspect, and \ding{56} indicates that this aspect is not covered. Multimodality  means that the benchmark has multiple languages, character images and mermaid drawings. Redundant Content indicates the noise content and background.}
\begin{adjustbox}{max width=\textwidth}
\begin{tabular}{lcccccccccc}
\toprule
\multirow{2}{*}{\textbf{Dataset}} 
& \multirow{2}{*}{\textbf{Linear}} 
& \multicolumn{3}{c}{\textbf{Non-linear}} 
& \multirow{2}{*}{\textbf{Table}} 
& \multirow{2}{*}{\textbf{\shortstack{Multimodality}}} 
& \multirow{2}{*}{\textbf{\shortstack{Reference \\Code}}} 
& \multirow{2}{*}{\textbf{\shortstack{Redundant \\Content} }} 
& \multirow{2}{*}{\textbf{AvgLen}} \\
& & \textbf{QP \& SOCP} 
& \textbf{HP \& FP} 
& \textbf{ELP} & & & & & & \\
\midrule
NLP4LP              & \ding{52} & \textcolor{red}{\ding{56}} & \textcolor{red}{\ding{56}} & \textcolor{red}{\ding{56}} & \textcolor{red}{\ding{56}} & \textcolor{red}{\ding{56}} & \textcolor{red}{\ding{56}} & \textcolor{red}{\ding{56}} & 536 \\
NL4OPT              & \ding{52} & \textcolor{red}{\ding{56}} & \textcolor{red}{\ding{56}} & \textcolor{red}{\ding{56}} & \textcolor{red}{\ding{56}} & \textcolor{red}{\ding{56}} & \textcolor{red}{\ding{56}} & \textcolor{red}{\ding{56}} & 534 \\
ComplexOR           & \ding{52} & \textcolor{red}{\ding{56}} & \textcolor{red}{\ding{56}} & \textcolor{red}{\ding{56}} & \textcolor{red}{\ding{56}} & \textcolor{red}{\ding{56}} & \ding{52} & \textcolor{red}{\ding{56}} & 1273 \\
MAMO (H)            & \ding{52} & \textcolor{red}{\ding{56}} & \textcolor{red}{\ding{56}} & \textcolor{red}{\ding{56}} & \textcolor{red}{\ding{56}} & \textcolor{red}{\ding{56}} & \textcolor{red}{\ding{56}} & \textcolor{red}{\ding{56}} & 1724 \\
ORllmAgent          & \ding{52} & \textcolor{red}{\ding{56}} & \textcolor{red}{\ding{56}} & \textcolor{red}{\ding{56}} & \ding{52} & \textcolor{red}{\ding{56}} & \textcolor{red}{\ding{56}} & \textcolor{red}{\ding{56}} & 1220 \\
OptiBench           & \ding{52} & \ding{52} & \textcolor{red}{\ding{56}} & \textcolor{red}{\ding{56}} & \ding{52} & \textcolor{red}{\ding{56}} & \textcolor{red}{\ding{56}} & \textcolor{red}{\ding{56}} & 686 \\
IndustryOR          & \ding{52} & \textcolor{red}{\ding{56}} & \textcolor{red}{\ding{56}} & \textcolor{red}{\ding{56}} & \ding{52} & \textcolor{red}{\ding{56}} & \textcolor{red}{\ding{56}} & \textcolor{red}{\ding{56}} & 1185 \\
OPTMATH             & \ding{52} & \ding{52} & \textcolor{red}{\ding{56}} & \textcolor{red}{\ding{56}} & \ding{52} & \textcolor{red}{\ding{56}} & \textcolor{red}{\ding{56}} & \textcolor{red}{\ding{56}} & 2974 \\
\textbf{NEXTOR}     & \ding{52} & \ding{52} & \ding{52} & \ding{52} & \ding{52} & \ding{52} & \ding{52} & \ding{52} & \textbf{3144} \\
\bottomrule
% \multicolumn{11}{l}{{Only NEXTOR (L) covers full-spectrum non-linear problems and long-form realistic constraints with redundancy, code examples, and visual noise.}}
\end{tabular}
\end{adjustbox}
\label{table:benchmark_comparison}
\end{table*}

The comparison between its data and other data is shown in Table ~\ref{table:benchmark_comparison}

\subsection{NXETOR’s Statistics.}

As shown in Figure~\ref{fig:benchmark_Statistics}(a), we provide statistics for 10 benchmarks' average length. Figure~\ref{fig:benchmark_Statistics}(b.1) further illustrate the distributions of 8 distinct problem types.  Different from other benchmarks that only involve QP or SCOP problems, our benchmark involves various complex types of nonlinear problems.This is because complex nonlinear functions are often encountered in practical problems.,  such as in finance fields, Cobb-Douglas production functions $Q(L,K)=A_t L^{\alpha} K^{\beta}\mu$ introduce power functions with diminishing marginal returns. Figure~\ref{fig:benchmark_Statistics}(b.2-3) further illustrate question length and variable numbers distribution. It is evident that instances in our long and complex category feature a significantly higher number of variables, with text lengths concentrated around 3,500 tokens, positioningNXETOR's average problem length among all benchmarks. Overall,NXETOR demonstrates substantial diversity in terms of problem types, variable counts, and text lengths.

\subsection{NEXTOR data case.} 

The following is aNXETOR data example, including index, question, and answer.

\begin{lstlisting}[style=mdblock]
"1": {
"index": 1,
"question": 
    "Located under the shade of sycamore trees in the old
    city, Changning District People's Hospital has been a
    trusted medical center for surrounding residents for 
    decades. With the aging of the population and the 
    opening of the two-child policy, the number of 
    outpatient    visits and hospitalization needs have
    increased year by year, especially in the pediatric
    and emergency departments, which are often 
    overcrowded. Head nurse Lin Mei has worked in this
    hospital for 15 years. She arrives at work at five in
    the morning every day to check the shift schedule and
    coordinate emergencies - for example, Xiao Li, who
    worked the night shift yesterday, suddenly had a fever
    , and twins were born in the obstetrics department in
    the early morning. She knows that scheduling is not
    just a table, but also about patient safety and the
    morale of the nursing team: newcomers need to be
    mentored by old staff, pregnant women cannot work
    night shifts, and nurses who work around the clock
    are prone to mistakes... But the hospital budget is
    limited, and human resource costs must be carefully
    calculated. Under the premise of meeting the demand
    for nurses in each time period, design an efficient
    and humane scheduling plan to avoid overwork or waste
    of manpower, and reserve flexible space for emergencies.

    People's Hospital is a district-level hospital in
    Changning District. The number of nurses on duty
    required in each time period of the hospital every
    day is shown in Table C-2. 


    Table C-2
    | Time period | 6:00-10:00 | 10:00-14:00 | 14:00-18:00 | 18:00-22:00 | 22:00-6:00 (next day) |
    | -------- | ---------- | ----------- | ----------- | ----------- | ---------------- |
    | Number of nurses required | 18 | 20 | 19 | 17 | 12 |


    The nurses in this hospital work in five shifts, each 8
    hours. The specific working hours are 2:00-10:00 for the
    first shift, 6:00-14:00 for the second shift, 
    10:00-18:00 for the third shift, 14:00-22:00 for the
    fourth shift, and 18:00-2:00 (next day). Each nurse
    works 5 shifts per week and is assigned to different
    days. There is a chief nurse who is responsible for the
    nurses' duty arrangements and duty plans, which should
    be relatively economical in terms of personnel or
    economy, and as reasonable as possible. It is the duty
    plan:
    
    **Plan**: Each nurse works for 5 consecutive days, rests
    for 2 days, and is arranged from the first shift to the
    fifth shift from the first day of work. For example, if
    a nurse starts working on Monday, she will work the
    first shift on Monday, the second shift on Tuesday...
    the fifth shift on Friday; if another nurse starts
    working on Wednesday, she will work the first shift on
    Wednesday, the second shift on Thursday... the fifth
    shift on Sunday, and so on.
    
    Establish a linear programming model to minimize the
    number of nurses on duty. You only need to give the
    minimum number of nurses. ",

"answer": 70.0
}
\end{lstlisting}

\subsection{NXETOR Guiding Principles.}
\label{app:guiding_principles}
The content ofNXETOR is based on 4 core principles, each of which corresponds to the questions we raised in \hyperref[sec:introduction]{\textbf{Introduction}}. \textbf{(a) Coverage Extension} aims to transcend the limitations of traditional benchmarks by incorporating a wider array of nonlinear and complex problem types. \textbf{(b) Complexity Elevation} seeks to simulate the depth of real-world decision-making by increasing the intrinsic scale of variables and constraints. \textbf{(c) Realism Augmentation} moves beyond pristine mathematical descriptions by integrating redundancy, unstructured data, and multi-modal information characteristic of real-world scenarios. Lastly, \textbf{(d) Verifiability Guarantee} ensures that every instance is solvable and accompanied by a ground-truth solution and reference solver code.

\begin{figure*}[!htbp]
  \centering
  \includegraphics[width=0.95\linewidth]
  {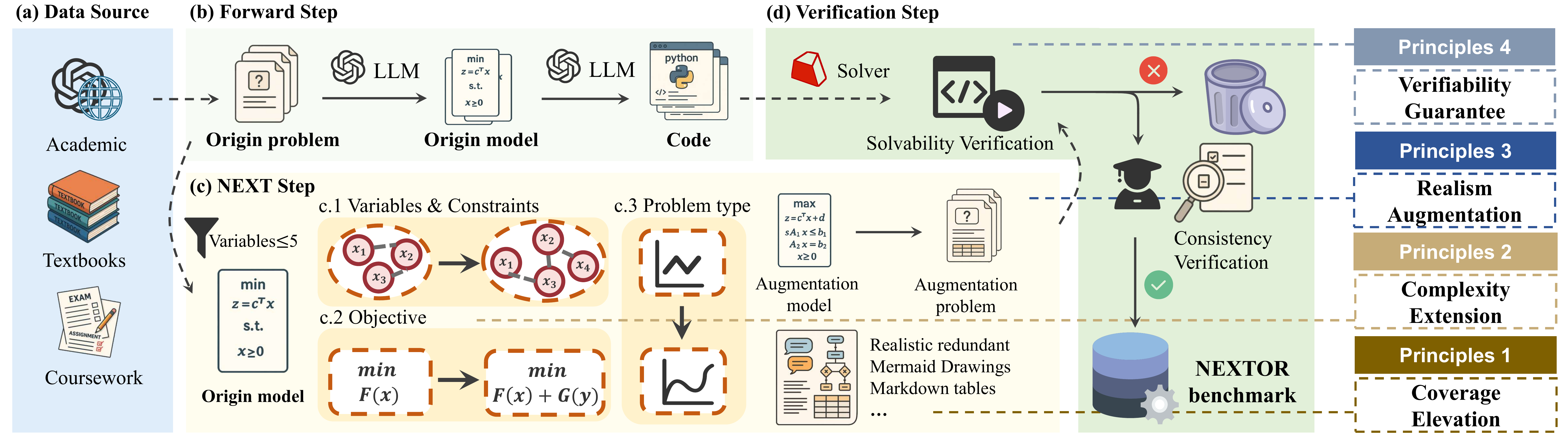}
  \caption{NEXTOR Synthesis Method. This synthesis method first (a) collects data from three primary channels, and then proceeds through three core steps: (b) the Forward Step, (c) the NEXT Step, and (d) the Verification Step, all based on four Guiding Principles.}
  \label{fig:NETA_Framework}
\end{figure*}

\subsection{NXETOR Benchmark and Synthesis Method}
\label{sec:NEXTOR}

To facilitate the evaluation of models on complex OR problems and to assess the effectiveness of NEDTree, we have constructed a new benchmark, NXETOR, which addresses the limitations of existing benchmarks in capturing realistic scenarios. This section introduces the data generation method, and benchmark comparisons are provided in Section \ref{sec:setup}. The overall benchmark synthesis method, illustrated in Figure~\ref{fig:NETA_Framework}, comprises three main stages: the Forward Step, the NEXT Step, and the Verification Step. This process is based on four guiding principles, the details of which are provided in Appendix \ref{app:guiding_principles}.

\subsubsection{Data Source.}
\label{sec:data_source.} We source our raw data through three primary channels as shown in Figure~\ref{fig:NETA_Framework}(a): (a) searching for source materials using OpenAI's Deep Research, (b) collecting problems from established textbooks~\cite{Hu2018_OR_Tutorial,Hu2019_OR_Exercises, Wolsey2020_IntegerProgramming}, and (c) collecting problems from university course assignments and exams. We will remove all names and add pseudonyms to avoid personally identifying information or offensive content. Furthermore, we only collect publicly available data from the internet, and there are no issues regarding data privacy.

\subsubsection{Forward step.}
\label{sec:forward_Step}
The main objective of the Forward Step is to solve and standardize these problem seeds. This process involves: (a) formally defining the decision variables, objective function, and constraints; (b) writing standard Python solver code (primarily using Gurobi); and (c) executing the code to obtain a deterministic optimal solution, which serves as the ground truth (GT). The output of this stage is an internal ``seed dataset'', where each instance comprises a natural language description, a structured mathematical model, standard solver code, and the GT answer.

\subsubsection{NEXT Step.}
\label{sec:NEXT_Step}
The NEXT step is the core of the method, where problem seeds are enhanced for LLM evaluation. In this paper, we select problems with less than 5 variables for augmentation, as more than 80\% of the problems have less than 5 variables in representative OPTIBench dataset\cite{yang2024optibench}. This stage proceeds along 2 main dimensions in Figure ~\ref{fig:NETA_Framework}(c). The first is Complexity Enhancement, aimed at increasing the intrinsic mathematical difficulty of the problems, shown in Figure \ref{fig:NETA_Framework}(c.1-c.3). This is achieved by expanding the number of decision variables and constraints, introducing reasonable modifications or perturbations to the objective function, and promoting basic problem types to more advanced forms (e.g., LP to FP). The second dimension is Realism Augmentation, which focuses on simulating real-world information environments. We strategically inject redundant information (e.g., irrelevant background stories, distracting data points) and fuse multi-modal information (e.g., embed key parameters in Markdown tables or Mermaid charts) into the problem descriptions. These augmentations significantly raise the requirements for information extraction, filtering, and noise immunity capabilities.

% \paragraph{Quality Assurance Mechanism}
\subsubsection{Verification Step.}
To ensure the validity of the NXETOR benchmark, every augmented problem from the NEXT step must pass a strict quality assurance. First is Solvability Verification, where we write code manually to solve the augmented problem, verifying that it remains solvable and has a deterministic solution. Second is Consistency Verification, where OR experts conduct a comprehensive review to assess the logical coherence of the problem description with the final answer. Only problems that successfully pass entire verification steps are incorporated into the final NXETOR.

\section{Other results}
\begin{table*}[h]
\centering
\caption{Comparison of Efficiency between CoE and NED-Tree.}
\label{tab:efficiency_comparison}
\resizebox{\linewidth}{!}{
\begin{tabular}{l l l l l}
\hline
Dataset & Method & Accuracy & Avg. time (s) & Avg. tokens \\
\hline
NL4OPT & CoE & 64.20\% $\pm$ 0.003295 & 303.67 & 6965.32 \\
NL4OPT & NED-Tree & 94.35\% $\pm$ 0.001705 & 35.21 & 4968.28 \\
NEXTOR-Linear & CoE & 49.38\% $\pm$ 0.0007685 & 175.579 & 17795.71 \\
NEXTOR-Linear & NED-Tree & 60.52\% $\pm$ 0.0006925 & 105.88 & 19871.77 \\
NEXTOR-Nonlinear & CoE & 39.62\% $\pm$ 0.0026039 & 309.73 & 13094.94 \\
NEXTOR-Nonlinear & NED-Tree & 92.11\% $\pm$ 0.0019621 & 75.46 & 14571.21 \\
\hline
\end{tabular}
}
\end{table*}

\textbf{Our Framework Demonstrates High Efficiency and Broad Generalization.}
We conducted a detailed comparison of NED-Tree and the prompt-based method Chain-of-Experts (CoE) regarding inference time and token consumption, as shown in Table \ref{tab:efficiency_comparison}. On the NL4OPT dataset, the average inference time of NED-Tree is only 35.21 seconds, demonstrating faster inference speed. Even in the more complex NEXTOR-Linear and NEXTOR-Nonlinear tasks, NED-Tree saves approximately 40\% and 43\% of the time, respectively. This is primarily attributed to the structured generation paradigm of NED-Tree, which avoids the time overhead caused by multi-agent repeated dialogue and lengthy reasoning inherent in the CoE framework.

\begin{table}[h]
\centering
\caption{Performance Comparison with Different Solvers.}
\label{tab:solver_comparison}
\begin{adjustbox}{max width=0.48\textwidth}
\begin{tabular}{l l l}
\hline
Solver & NL4OPT & NEXTOR \\
\hline
Gurobi & 94.35\% $\pm$ 0.0006529 & 76.31\% $\pm$ 0.0005526 \\
COPT & 94.29\% $\pm$ 0.0007132 & 76.22\% $\pm$ 0.0006634 \\
\hline
\end{tabular}
\end{adjustbox}
\end{table}

Secondly, existing fine-tuned models often overfit to the API syntax of specific solvers (such as Gurobi), leading to poor transferability. In contrast, NED-Tree possesses native solver agnosticism by constructing intermediate representations (Definitions and Final Linear Form). To verify this, we compared the performance of two solvers, Gurobi and COPT, by replacing only the back-end knowledge injection module without altering the upstream extraction and decomposition logic. As shown in Table \ref{tab:solver_comparison}, the experimental results indicate that on the NL4OPT dataset, the accuracy difference between using Gurobi (94.35\%) and COPT (94.29\%) is only 0.06\%; on the more challenging NEXTOR dataset, the difference is also merely 0.09\% (76.31\% vs 76.22\%). This result confirms that users can switch solvers at zero cost according to actual needs, greatly enhancing the potential for the framework's deployment in real-world industrial scenarios.

\section{Full Case Study}
\label{app:Full_case_study}
\begin{figure*}
    \centering
    \includegraphics[width=1\linewidth]{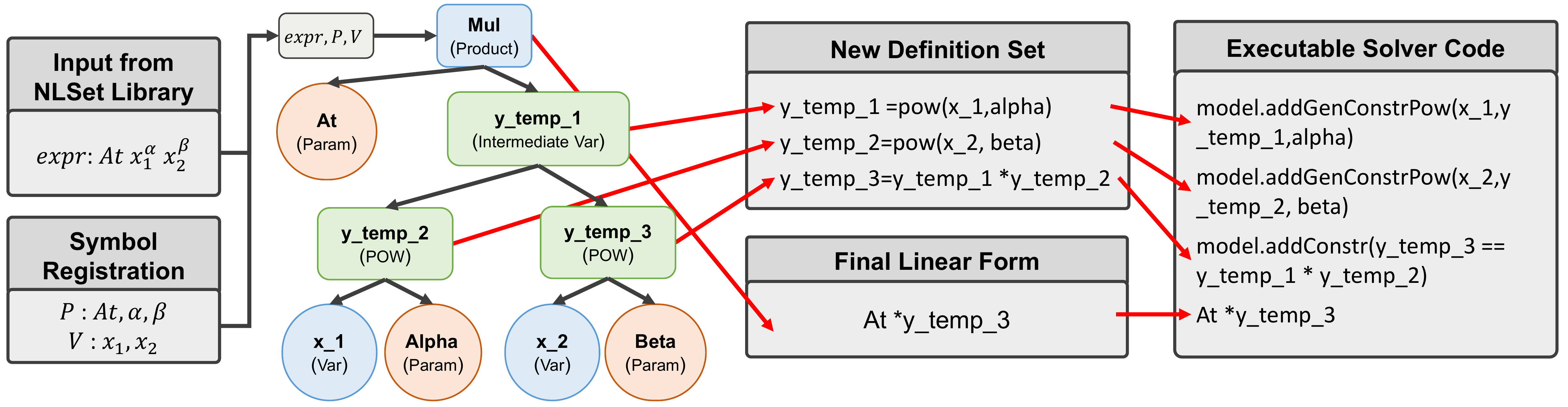}
    \caption{Case Study for the nonlinear term $At x_1^{\alpha}x_2^{\beta}$}
    \label{fig:case_study}
\end{figure*}
\subsection{Case Study}
\label{sec:case_study}

We present a case study on the decomposition of a nonlinear term common in industrial constraints: $At \cdot x_1^\alpha \cdot x_2^\beta$, as illustrated in Figure \ref{fig:case_study}. Here, the input is the nonlinear term $At \cdot x_1^\alpha \cdot x_2^\beta$, declaring $x_1, x_2$ as variables and $At, \alpha, \beta$ as parameters. NED-Tree first performs symbol registration, identifying $x_1, x_2$ as variables and $At, \alpha, \beta$ as parameters. Then, it recursively constructs the NED tree: (a) Leaf decomposition: The power terms $x_1^\alpha$ and $x_2^\beta$ are identified as atomic nonlinear units. The model generates definitions `y\_temp\_1 = pow(x\_1, alpha)` and `y\_temp\_2 = pow(x\_2, beta)`; (b) Internal node construction: The product of these two terms is handled by introducing an auxiliary variable `y\_temp\_3 = y\_temp\_1 * y\_temp\_2`; (c) Final term: The final term is linearly represented as `At * y\_temp\_3`. This structured decomposition allows the code agent to strictly call `model.addGenConstrPow` to map the leaves and standard quadratic constraints, ensuring executability.

\section{Potential Risks}
While NED-Tree significantly improves the reliability of automated modeling, applying Large Language Models to Operations Research entails inherent risks. First, optimization models are often deployed in high-stakes critical infrastructure (e.g., energy grids, supply chains). Despite our framework's alignment capabilities, the probabilistic nature of LLMs may still introduce subtle semantic errors or hallucinations that, while syntactically correct, could lead to costly or unsafe operational decisions. Second, the "executability" of the generated code might induce automation bias, where non-expert users blindly trust the solver's output without rigorously verifying the underlying mathematical formulation. Therefore, we strongly advocate for human-in-the-loop verification when deploying such systems in real-world industrial environments.

\section{Ai Assistants In Research Or Writing}
We used Gemini 3 pro for generating and organizing the initial draft of our paper, as well as for polishing it. We also referenced some images generated by Gemini as draft images for the paper, but we redraw all the figures.

\section{Extraction Implementation Details. }
\subsection{LLM Agent for birdeye Extraction} The following is the basic prompt template for Birdeye Extraction in sentence-by-sentence extraction strategy.

\textbf{System prompt template:}

\begin{lstlisting}[style=mdblock]
You are an expert in mathematical modeling and a professor of optimization at a top university. We will describe an optimization problem for you. Regarding this combinatorial optimization problem, please complete the following tasks:
    1. **Sentence Scanning**: Start by providing the original sentence number and content, and then scan sentence by sentence. EITHER extract it into one or more
    constraints information OR just mark it as "No Values".
    
    2. **Extract Parameters**: Extract all parameters from the following paragraphs and tables, making sure that no elements in any sentence are omitted. Specifically, you need to give a **parameter list**, provide the names of all parameters, and must indicate the parameter type (integer/float/list/tuple) and give specific values.
        Note 1. The "Value" of the list/tuple type are defined by using the python format, and should not be string. Example, a list can use ["S", "V"] or a tuple type can use {"A": 450, "B": 400, "C": 300} and so on.
        
        Note 2. If the problem description contains **table** data (usually in markdown format), please strictly convert the table data into the form of a list or tuple in the python language. You must strictly refer to the data I provide and do not make up your own data. In the end, you must also extract the table data and name it Table_1_XXX, Table_2_XXX, and so on.
        
        Note 3. The step you are processing now is only used to find parameters with specific values, and you do not need to consider decision variables or other constraints!
    3. **Output** as follows:
        1.Sentence Scanning Result
        sentence 1:<sentence 1> -> <Constraint Scanning result description OR 'No Values'>,
        sentence 2:<sentence 2> -> <Constraint Scanning result description OR 'No Values'>,
        ...
        2.Table Scanning Result
        table 1:<table_1_name> -> <Parameter Values(list/tuple)>,
        table 2:<table_2_name> -> <Parameter Values(list/tuple)>,
...
\end{lstlisting}

\textbf{User prompt template:}

\begin{lstlisting}[style=mdblock]
Here is the problem description:
    ________________________________________
    {user_question}
    Output the result mentioned above.
\end{lstlisting}

\textbf{User prompt template for format change:}

\begin{lstlisting}[style=mdblock]
Now, please convert all your analysis results(Sentence Parameters and Table Parameters) into **Parameters List**. Please adhere **strictly** to the following rules when generating the JSON field **"Value"**:
    1. Output must be **valid JSON**:  
    - All keys and string values in double quotes.  
    - No Python tuple syntax `(a, b)`.  
    - No objects with numeric or tuple keys.
    
    2.Value must follows these rules:  
    - If the original key is a **string**, keep the object structure.
    - If the original key is an **integer** (0,1,2,...), output a **one-dimensional** array. Element at index i corresponds to the value for key i.  
    - If the original key is an **integer pair** `[i,j]`, output a **two-dimensional square matrix**:  
     
     Now, use the following JSON object format (no additional text):  
        ```json
        {
            "Parameters_List": [
                {
                    "Name": "<Name of parameter1>",
                    "Type": "<integer/float/list/tuple>",
                    "Value": <Parameter Values, not string>,
                },
                ...
            ]
        }
\end{lstlisting}

% \begin{figure}[h]
%   \centering
%   \includegraphics[width=\linewidth]{fig/Appendix12.pdf}
%     \label{app:Parameters_Extraction}
%     \caption{Long-Search Agent(for Parameters Extraction).}
% \end{figure}
\subsection{LLM Agent for sentence-by-sentence extraction} The following is the basic prompt template for sentence-by-sentence extraction to get all Detailed OR elements.

\textbf{System prompt template:}

\begin{lstlisting}[style=mdblock]
You are an expert mathematical modeler and an optimization professor at a top university. We will give you a description of an optimization problem.
    Regarding this combinatorial optimization problem, please complete the following tasks:
    Extract all decision variables and constraints from the following paragraph, ensuring that no element from the sentences is overlooked.
    
    1. **Sentence Scanning**: Start by providing the original sentence number and content, and then scan sentence by sentence. EITHER extract it into one or more constraints information OR mark it as "No constraints".
    2. **Variable List**: Give Variables from constraints sentence, and point: Name(symbol) / Meaning / type:<integer type OR continuous type>" / Range of Values. 
    3. **Mapping Table**: In a Markdown table, precisely correspond the "Constraint Name <--> Mathematical Expression <--> Sentence Number."
    4. **Optimization Goal**: Provide the optimization objective (target or performance metric to be optimized).
    5. **Problem Type**: Determine whether the model is a MILP (Mixed Integer Linear Programming) problem or an NLP (Nonlinear Programming) problem, and select one of the two values. 
    
    Note:
        1. List all variables, including those introduced for linearizing absolute differences, such as $\delta^{+}$, $\delta^{-}$ (if such variables exist, list them, otherwise leave them out), and generate the corresponding linearization constraints.  For each original sentence, scan and check if keywords like "change," "difference," "increment," "decrement," "change amount," etc., are mentioned, and generate the corresponding linearization constraints.
        2. "not need" does not necessarily mean that there are no variables or constraints. If an object appears in sentences such as "does not need to increase" or "will not increase", it may be necessary to consider the situation where the variable will decrease.
        3. Do not ignore sentences starting with "In addition", "In addition to this", "By the way", etc., which may also contain information such as constraints or variables.
        4. If "all the sub-quotas" or "all types" are mentioned, then every category of situation must be considered.
    **Output** as follows:  
          1.Sentence_Scanning
            sentence 1:<sentence 1> -> <Constraint Scanning result description OR 'No constraints'>,
            sentence 2:<sentence 2> -> <Constraint Scanning result description OR 'No constraints'>,
            ...
          2.Variables_List
            Variable 1:...,
            Variable 2:...,
            ...
          3.Constraint_Table
            ["<Constraint 1 name>","<Mathematical expressions 1>","sentence numbers:<sentence numbers>"],
            ["<Constraint 2 name>","<Mathematical expressions 2>","sentence numbers:<sentence numbers>"],
            ...
          4.Objective
            <Objective sentence> and <Mathematical expressions>,
            ...
          5.Problem_Type
            point <'MILP' OR 'NLP'>, and give description...
\end{lstlisting}

\textbf{User prompt template:}

\begin{lstlisting}[style=mdblock]
Here is the problem description:
    ________________________________________
    {user_question}
    Output the lists(Sentence_Scanning, Variables_List, Constraint_Table,Objective, Problem_Type) mentioned above.
\end{lstlisting}

\textbf{User prompt template for format change:}

\begin{lstlisting}[style=mdblock]
Now, please convert all your analysis results into the following JSON object format (no additional text):  
    For origin sentence, just write the first few words and "...".
    ```json
    {
      "Sentence_Scanning": [
        ["1","<origin sentence 1...>","<Constraint Scanning result description OR 'No constraints'>"],
        ["2","<origin sentence 2...>","<Constraint Scanning result description OR 'No constraints'>"],
        ...
      ],
      "Variables_List": [
        {
          "symbol":     "<chosen mathematical symbol>",
          "Meaning":    "<parameter definition>",
          "Type":       "<BINARY / integer / continuous type>",
          "Range ":     "<Range of Values>"
        },
        ...
      ],
      "Constraint_Table":[
        ["<Constraint 1 name>","<Mathematical expressions 1>","sentence numbers:<sentence numbers>"],
        ["<Constraint 2 name>","<Mathematical expressions 2>","sentence numbers:<sentence numbers>"],
        ...
      ],
      "Objective": {
          "Objective_sentence":        "<Objective sentence>",
          "Mathematical_expressions":  "<Mathematical expressions>"
      },
      "Problem_Type": "<'MILP' OR 'NLP'>"
    }```
\end{lstlisting}

\end{document}